\documentclass[11pt,a4paper]{article}
\usepackage[hyperref]{emnlp-ijcnlp-2019}
\usepackage{times}
\usepackage{latexsym}
\usepackage{graphicx}
\usepackage{multirow}
\usepackage{float}
\usepackage{amssymb}
\usepackage{url}
\usepackage{booktabs}

\usepackage{helvet} 
\usepackage{courier} 

\usepackage{color,soul}
\usepackage{amsmath,mathtools}
\usepackage{amsfonts}
\usepackage[capitalize]{cleveref}
\usepackage{multirow,multicol}
\usepackage{MnSymbol}
\usepackage{array}
\usepackage{subcaption}
\usepackage{pifont}
\newcommand{\cmark}{\ding{51}}%
\newcommand{\xmark}{\ding{55}}

\newcolumntype{L}[1]{>{\raggedright\let\newline\\\arraybackslash\hspace{0pt}}m{#1}}
\newcolumntype{C}[1]{>{\centering\let\newline\\\arraybackslash\hspace{0pt}}m{#1}}
\newcolumntype{R}[1]{>{\raggedleft\let\newline\\\arraybackslash\hspace{0pt}}m{#1}}

\aclfinalcopy 

\DeclareMathOperator*{\ReLU}{\text{ReLU}}
\DeclareMathOperator*{\softmax}{\text{softmax}}
\DeclareMathOperator*{\argmax}{\text{argmax}}
\newcommand\norm[1]{\left\lVert#1\right\rVert}

\makeatletter
\def\thanks#1{\protected@xdef\@thanks{\@thanks
        \protect\footnotetext{#1}}}
\makeatother

\title{DialogueGCN: A Graph Convolutional Neural Network for Emotion~Recognition~in~Conversation} 

\author{Deepanway Ghosal$^\dagger$, Navonil Majumder$^\ddagger$, Soujanya Poria$^{\dagger\ast}$\thanks{$^\ast$ Corresponding author},\\ \textbf{Niyati Chhaya$^\nabla$ and Alexander Gelbukh$^\ddagger$}\\
 $^\dagger$ Singapore University of Technology and Design, Singapore \\
 $^\ddagger$Instituto Polit\'ecnico Nacional, CIC, Mexico \\
 $^\nabla$ Adobe Research, India\\
\\ {\tt \{1004721@mymail.,sporia@\}sutd.edu.sg}, {\tt navo@nlp.cic.ipn.mx},\\ {\tt nchhaya@adobe.com}, {\tt gelbukh@gelbukh.com} 
}
 
\date{}

\begin{document}

\maketitle

\begin{abstract}

Emotion recognition in conversation (ERC) has received much attention, lately, from researchers due to its potential widespread applications in diverse areas, such as health-care, education, and human resources. In this paper, we present Dialogue Graph Convolutional Network (DialogueGCN), a graph neural network based approach to ERC. We leverage self and inter-speaker dependency of the interlocutors to model conversational context for emotion recognition. Through the graph network, DialogueGCN addresses context propagation issues present in the current RNN-based methods. We empirically show that this method alleviates such issues, while outperforming the current state of the art on a number of benchmark emotion classification datasets.
\end{abstract}

\section{Introduction}

\begin{figure}[t]
    \centering
    \includegraphics[width=\linewidth]{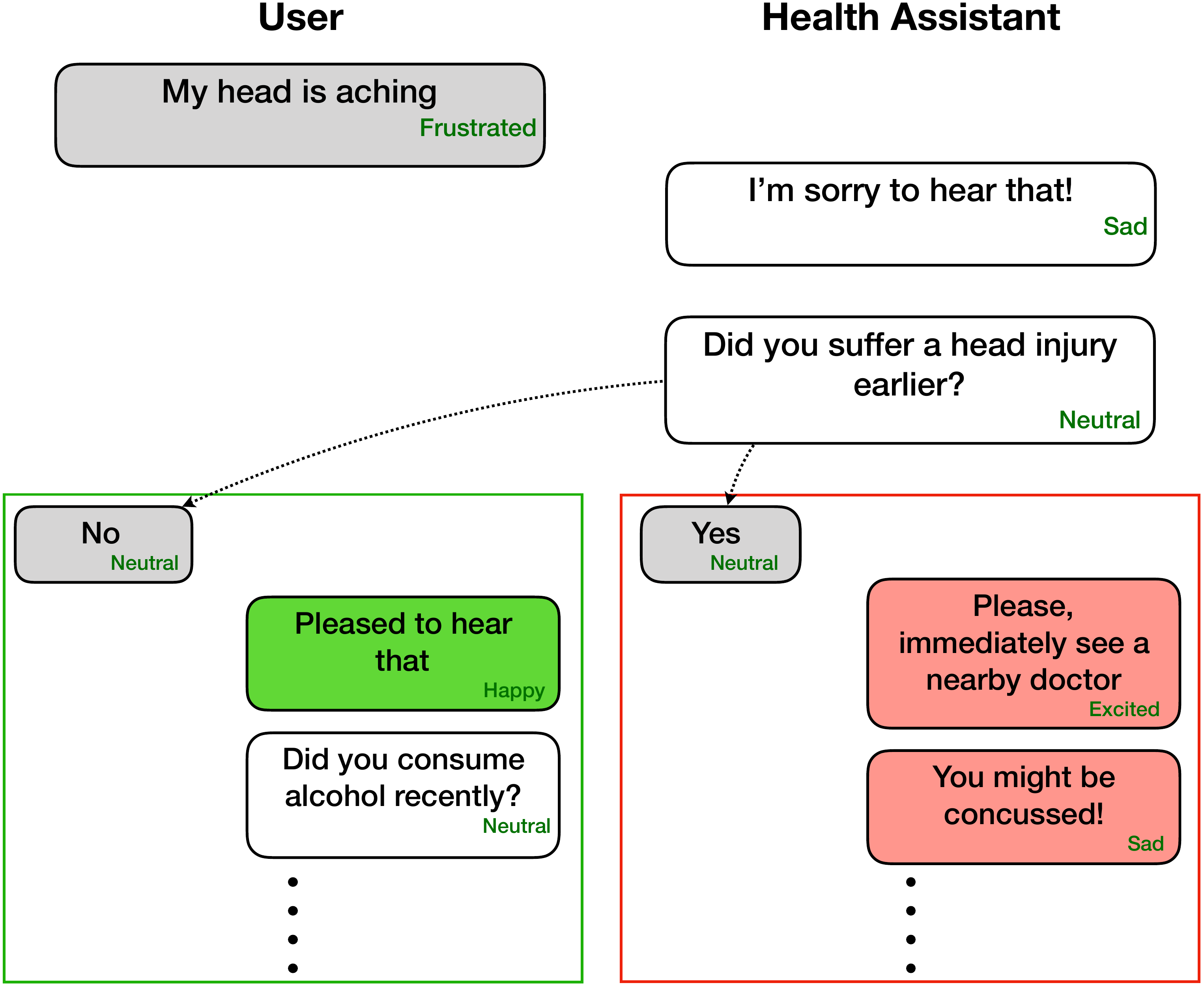}
    \caption{Illustration of an affective conversation where the emotion depends on the context. Health assistant understands affective state of the user in order to generate affective and empathetic responses.}
    \label{fig:affective-dialogue}
\end{figure}

Emotion recognition has remained an active research topic for decades \cite{dmello,iemocap,strapparava2010annotating}. However, the recent proliferation of open conversational data on social media platforms, such as Facebook, Twitter, Youtube, and Reddit, has warranted serious attention~\cite{poria2019emotion,dialoguernn,huang2019ana} from researchers towards emotion recognition in conversation (ERC). ERC is also undeniably important in affective dialogue systems (as shown in \cref{fig:affective-dialogue}) where bots understand users' emotions and sentiment to generate emotionally coherent and empathetic responses. 

\begin{figure}[t]
    \centering
    \includegraphics[width=\linewidth]{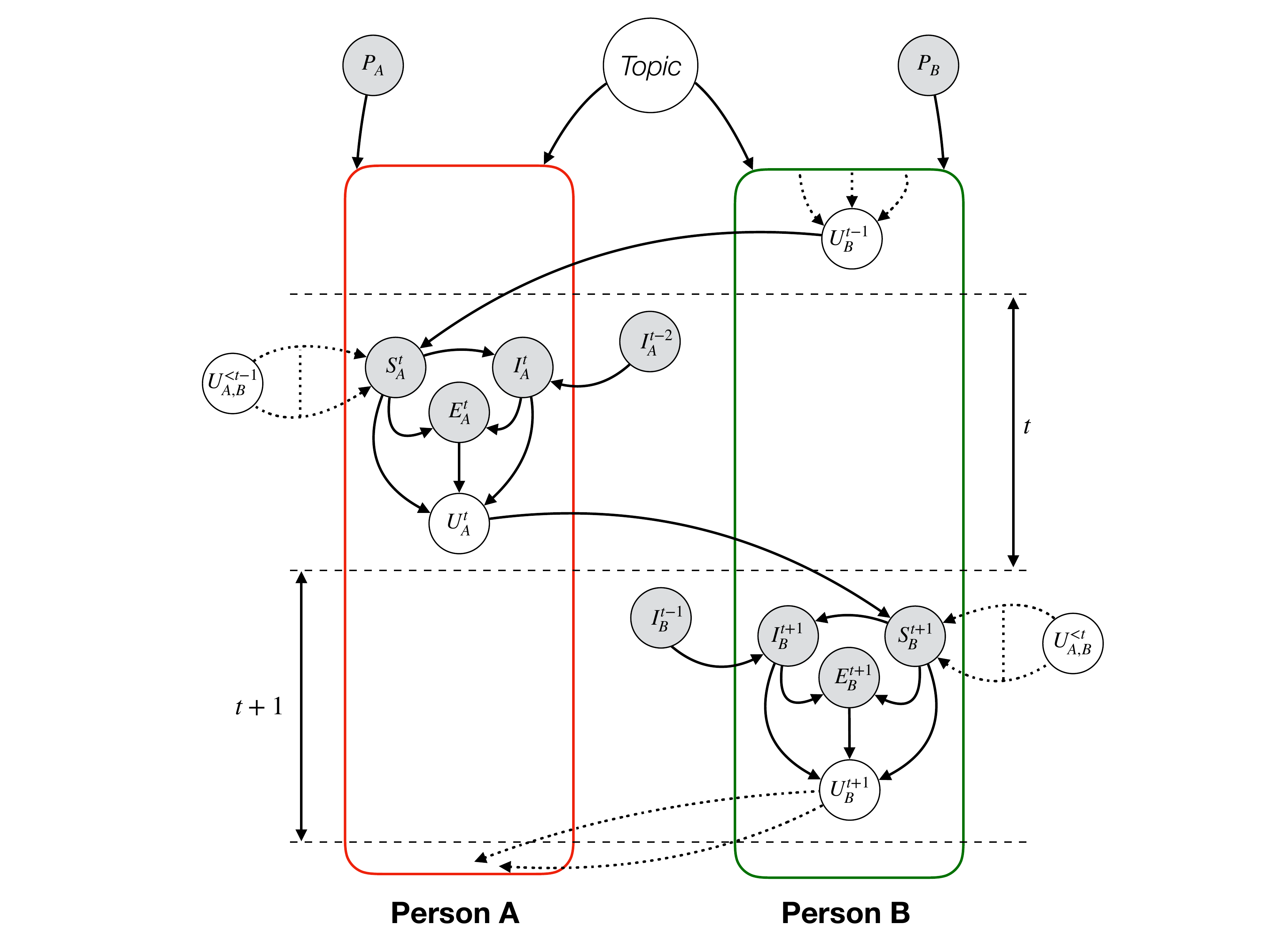}
    \caption{Interaction among different controlling variables during a dyadic conversation between persons A and B. Grey and white circles represent hidden and observed variables, respectively. $P$ represents personality, $U$ represents utterance, $S$ represents interlocutor state, $I$ represents interlocutor intent, $E$ represents emotion and \emph{Topic} represents topic of the conversation. This can easily be extended to multi-party conversations.}
    \label{fig:intent-modelling}
\end{figure}

Recent works on ERC process the constituent utterances of a dialogue in sequence, with a recurrent neural network (RNN). Such a scheme is illustrated in \cref{fig:intent-modelling}~\cite{poria2019emotion}, that relies on propagating contextual and sequential information to the utterances. Hence, we feed the conversation to a bidirectional gated recurrent unit (GRU)~\cite{gru}. However, like most of the current models, we also ignore intent modelling, topic, and personality due to lack of labelling on those aspects in the benchmark datasets. In theory, RNNs like long short-term memory (LSTM)~\cite{hochreiter1997long} and GRU should propagate long-term contextual information. However, in practice it is not always the case~\cite{bradbury2016quasi}. This affects the efficacy of RNN-based models in various tasks, including ERC.

To mitigate this issue, some variants of the state-of-the-art method, DialogueRNN~\cite{dialoguernn}, employ attention mechanism that pools information from entirety or part of the conversation per target utterance. However, this pooling mechanism does not consider speaker information of the utterances and the relative position of other utterances from the target utterance. Speaker information is necessary for modelling inter-speaker dependency, which enables the model to understand how a speaker coerces emotional change in other speakers. Similarly, by extension, intra-speaker or self-dependency aids the model with the understanding of emotional inertia of individual speakers, where the speakers resist the change of their own emotion against external influence. On the other hand, consideration of relative position of target and context utterances decides how past utterances influence future utterances and vice versa. While past utterances influencing future utterances is natural, the converse may help the model fill in some relevant missing information, that is part of the speaker's background knowledge but explicitly appears in the conversation in the future.
We leverage these two factors by modelling conversation using a directed graph. The nodes in the graph represent individual utterances. The edges between a pair of nodes/utterances represent the dependency between the speakers of those utterances, along with their relative positions in the conversation. By feeding this graph to a graph convolution network (GCN)~\cite{NIPS2016_6081}, consisting of two consecutive convolution operations, we propagate contextual information among distant utterances. We surmise that these representations hold richer context relevant to emotion than DialogueRNN. This is empirically shown in \cref{sec:results}.

The remainder of the paper is organized as follows --- \cref{sec:related-works} briefly discusses the relevant and related works on ERC; \cref{sec:method} elaborates the method; \cref{sec:experiments} lays out the experiments; \cref{sec:results} shows and interprets the experimental results; and finally, \cref{sec:conclusion} concludes the paper.

\section{Related Work}
\label{sec:related-works}
Emotion recognition in conversation is a popular research area in natural language processing~\cite{kratzwald2018decision, colneric2018emotion} because of its potential applications in a wide area of systems, including opinion mining, health-care, recommender systems, education, etc. 

However, emotion recognition in conversation has attracted attention from researchers only in the past few years due to the increase in availability of open-sourced conversational datasets~\cite{chen2018emotionlines,zhou2018emotional,poria2018meld}. A number of models has also been proposed for emotion recognition in multimodal data i.e. datasets with textual, acoustic and visual information. Some of the important works include ~\citep{poria-EtAl:2017:Long,chen2017multimodal,AAAI1817341,zadatt,hazarika2018icon,hazarika-EtAl:2018:N18-1}, where mainly deep learning-based techniques have been employed for emotion (and sentiment) recognition in conversation, in only textual and multimodal settings. The current state-of-the-art model in emotion recognition in conversation is~\cite{dialoguernn}, where authors introduced a party state and global state based recurrent model for modelling the emotional dynamics.

Graph neural networks have also been very popular recently and have been applied to semi-supervised learning, entity classification, link prediction, large scale knowledge base modelling, and a number of other problems ~\cite{kipf2016semi,schlichtkrull2018modeling,bruna2013spectral}. Early work on graph neural networks include ~\cite{scarselli2008graph}. Our graph model is closely related to the graph relational modelling work introduced in ~\cite{schlichtkrull2018modeling}.

\section{Methodology}
\label{sec:method}

One of the most prominent strategies for emotion recognition in conversations is contextual modelling. We identify two major types of context in ERC -- sequential context and speaker-level context. Following \citet{poria-EtAl:2017:Long}, we model these two types of context through the neighbouring utterances, per target utterance.

Computational modeling of context should also consider emotional dynamics of the interlocutors in a conversation. Emotional dynamics is typically subjected to two major factors in both dyadic and multiparty conversational systems --- inter-speaker dependency and self-dependency. Inter-speaker dependency refers to the emotional influence that counterparts produce in a speaker. This dependency is closely related to the fact that speakers tend to mirror their counterparts to build rapport during the course of a dialogue \cite{navarretta2016mirroring}. However, it must be taken into account, that not all participants are going to affect the speaker in identical way. Each participant 
generally affects each
other participants in unique ways. In contrast, self-dependency, or emotional inertia, deals with the aspect of emotional influence that speakers have on themselves during conversations. Participants in a conversation are likely to stick to their own emotional state due to their emotional inertia, unless the counterparts invoke a change. Thus, there is always a major interplay between the inter-speaker dependency and self-dependency with respect to the emotional dynamics 
in the conversation.

We surmise that combining these two distinct yet related contextual information schemes (sequential encoding and speaker level encoding) would create enhanced context representation leading to better understanding of emotional dynamics in conversational systems.
\subsection{Problem Definition}
\label{sec:problem-definition}
Let there be $M$ speakers/parties $p_1,p_2,\dots,p_M$ 
in a conversation. 
The task is to predict the emotion labels (\textit{happy}, \textit{sad}, \textit{neutral}, \textit{angry}, \textit{excited}, \textit{frustrated}, \textit{disgust}, and \textit{fear}) of the constituent utterances $u_1,u_2,\dots,u_N$, where utterance $u_i$ is uttered by speaker $p_{s(u_i)}$,
while $s$ being the mapping between utterance and index of its corresponding speaker. 
We also represent $u_i\in \mathbb{R}^{D_m}$ as the feature representation of the utterance,
obtained using the feature extraction process 
described below.



\subsection{Context Independent Utterance-Level Feature Extraction}
\label{sec:text-feat-extr}
A convolutional neural network~\cite{kim2014convolutional} is used to extract textual features from the transcript of the utterances. 
We use a single convolutional layer followed by max-pooling and a fully connected layer to obtain 
the feature representations for the utterances. The input to this network is the 300 dimensional pretrained 840B GloVe vectors~\cite{pennington2014glove}. We use filters of size 3, 4 and 5 with 50 feature maps in each. The convoluted features are then max-pooled with a window size of 2 followed by the ReLU activation~\cite{nair2010rectified}.
These are then concatenated and fed to a $100$ dimensional fully connected layer, whose activations form the representation of the utterance. 
This network is trained at utterance level with the emotion labels.





\subsection{Model}
\begin{figure*}[t]
    \centering
    \includegraphics[width=\textwidth]{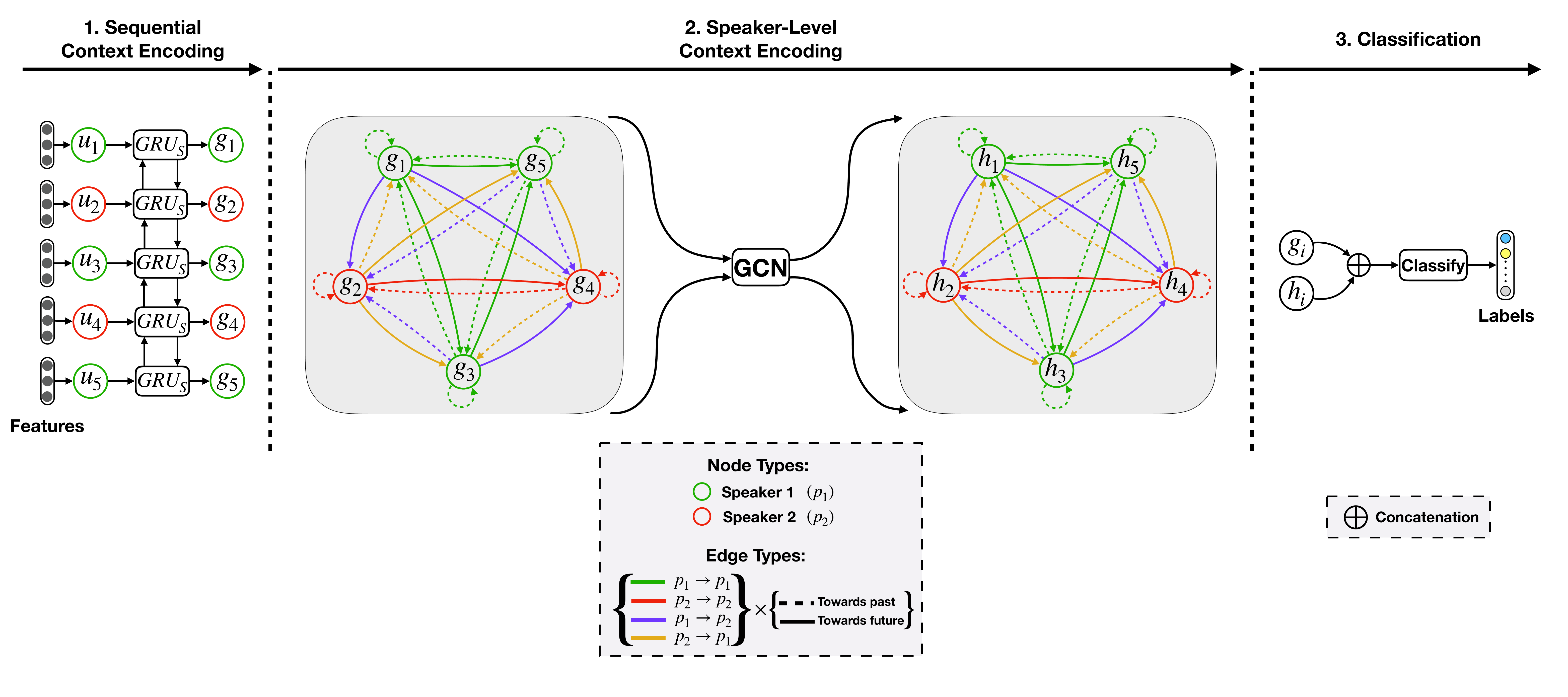}
    \caption{Overview of DialogueGCN, congruent to the illustration in \cref{example}.}
    \label{fig:model}
\end{figure*}

We now present our Dialogue Graph Convolutional Network (DialogueGCN\footnote{Implementation available at \url{https://github.com/SenticNet/conv-emotion}}) framework for emotion recognition in conversational setups. DialogueGCN consists of three integral components --- Sequential Context Encoder, Speaker-Level Context Encoder, and Emotion Classifier. An overall architecture of the proposed framework is illustrated in \cref{fig:model}.  

\subsubsection{Sequential Context Encoder} 
Since, conversations are sequential by nature, contextual information flows along that sequence. We feed the conversation to a bidirectional gated recurrent unit (GRU) to capture this contextual information:
 $g_i$ = $\overleftrightarrow{GRU_{\mathcal{S}}}(g_{i(+, -)1}, u_i)$, \text{for }i = 1, 2,\dots, N, where $u_i$ and $g_i$ are context-independent and sequential context-aware utterance representations, respectively.

Since, the utterances are encoded irrespective of its speaker, this initial encoding scheme is speaker agnostic, as opposed to the state of the art, DialogueRNN~\cite{dialoguernn}.

\subsubsection{Speaker-Level Context Encoder}
\label{sec:spealer-level}
We propose the Speaker-Level Context Encoder module in the form of a graphical network to capture speaker dependent contextual information in a conversation. Effectively modelling speaker level context requires capturing the inter-dependency and self-dependency among participants.
We design a directed graph from the sequentially encoded utterances to capture this interaction between the participants. Furthermore, we propose a local neighbourhood based convolutional feature transformation process to create the enriched speaker-level contextually encoded features. The framework is detailed here. 

First, we introduce the following notation: a conversation having $N$ utterances is represented as a directed graph $\mathcal{G = (V, E, R, W)}$, with vertices/nodes $v_{i} \in \mathcal{V}$, labeled edges (relations) 
$r_{ij} \in \mathcal{E}$ where $r \in \mathcal{R}$ is the relation type of the edge between $v_i$ and $v_j$ 
and $\alpha_{ij}$ is the weight of the labeled edge $r_{ij}$, with $0\leqslant \alpha_{ij} \leqslant 1$, where $\alpha_{ij} \in \mathcal{W}$ and $i, j \in [1,2,...,N]$. 

\paragraph{\textbf{Graph Construction:}} The graph is constructed from the utterances in the following way,

  \textbf{Vertices:} Each utterance in the conversation is represented as a vertex $v_i\in \mathcal{V}$ in $\mathcal{G}$. 
  Each vertex $v_i$ is initialized with the corresponding sequentially encoded feature vector $g_i$, for all $i \in [1,2,...,N]$. 
  We denote this vector as the vertex feature. Vertex features are subject to change downstream, when the neighbourhood based transformation process is applied to encode speaker-level context.
  
  \textbf{Edges:} Construction of the edges $\mathcal{E}$ depends on the context to be modeled. 
  For instance, if we hypothesize that each utterance (vertex) is contextually dependent on all the other utterances in a conversation (when encoding speaker level information), then a fully connected graph would be constructed. That is each vertex is connected to all the other vertices (including itself) with an edge. 
  However, this results in $O(N^2)$ number of edges, which is computationally very expensive for graphs with large number of vertices. 
  A more practical solution is to construct the edges by keeping a past context window size of $p$ and a future context window size of $f$. 
  In this scenario, each utterance vertex $v_i$ has an edge with the immediate $p$ utterances of the past: $v_{i-1},v_{i-2},..v_{i-p}$, $f$ utterances of the future: $v_{i+1},v_{i+2},..v_{i+f}$ and itself: $v_i$. For all our experiments in this paper, we consider a past context window size of 10 and future context window size of 10.
  
  As the graph is directed, two vertices can have edges in both directions with different relations. 
  
  \textbf{Edge Weights:} 
  The edge weights are set using a similarity based attention module. The attention function is computed in a way such that, for each vertex, the incoming set of edges has a sum total weight of 1. Considering a past context window size of $p$ and a future context window size of $f$, the weights are calculated as follows,
  \begin{equation}
  \begin{split}
    \alpha_{ij} & = \softmax(g_{i}^{T}W_{e}[g_{i-p},\dots,g_{i+f}])\label{eq:edge_weights}, \\
    & \text{for }j = i-p,\dots, i+f.
  \end{split}
  \end{equation}
  This ensures that, vertex $v_{i}$ which has incoming edges with vertices $v_{i-p},\dots,v_{i+f}$ (as speaker-level context) receives a total weight contribution of 1.

  \textbf{Relations:} The relation $r$ of an edge $r_{ij}$ is set depending upon two aspects:
  
      \textbf{\textit{Speaker dependency --- }} The relation depends on both the speakers of the constituting vertices: $p_{s(u_i)}$ (speaker of $v_i$) and $p_{s(u_j)}$ (speaker of $v_j$).
      
      \textbf{\textit{Temporal dependency --- }} The relation also depends upon the relative position of occurrence of $u_i$ and $u_j$ in the conversation: whether $u_i$ is uttered before $u_j$ or after.
  If there are $M$ distinct speakers in a conversation, there can be a maximum of $M$ (speaker of $u_i$) $* M$ (speaker of $u_j$) $ * 2$ ($u_i$ occurs before $u_j$ or after) $ = 2M^2$ distinct relation types $r$ in the graph $\mathcal G$.
  
  Each speaker in a conversation is uniquely affected by each other speaker, hence we hypothesize that explicit declaration of such relational edges in the graph would help in capturing the inter-dependency and self-dependency among the speakers in play, which in succession would facilitate speaker-level context encoding. 
  
  As an illustration, let two parties $p_1, p_2$ participate in a dyadic conversation having 5 utterances, where $u_1, u_3, u_5$ are uttered by $p_1$ and $u_2, u_4$ are uttered by $p_2$. 
  If we consider a fully connected graph, the edges and relations will be constructed as shown in \cref{example}.
\subparagraph{Feature Transformation:}
We now describe the methodology to transform the sequentially encoded features using the graph network. 
The vertex feature vectors ($g_i$) are initially speaker independent and thereafter transformed into a speaker dependent feature vector using a two-step graph convolution process. 
Both of these transformations can be understood as special cases of a basic differentiable message passing method~\cite{gilmer2017neural}.

In the first step, a new feature vector $h_i^{(1)}$ is computed for vertex $v_i$ by aggregating local neighbourhood information (in this case neighbour utterances specified by the past and future context window size) using the relation specific transformation inspired from \cite{schlichtkrull2018modeling}:
\begin{equation}
\begin{split}
h_{i}^{(1)}  & = \sigma (\sum\limits_{r \in \mathcal{R}}^{} \sum\limits_{j \in N_{i}^{r}}^{} \frac{\alpha_{ij}}{c_{i,r}}W_{r}^{(1)} g_{j} + \alpha_{ii}W_{0}^{(1)}g_{i}), \label{eq-ref1}\\
& \text{for }i = 1, 2,\dots, N,
\end{split}
\end{equation}
where, $\alpha_{ij}$ and $\alpha_{ii}$ are the edge weights, $N_{i}^{r}$ denotes the neighbouring indices of vertex $i$ under relation $r \in \mathcal{R}$. 
$c_{i,r}$ is a problem specific normalization constant which either can be set in advance, such that, $c_{i,r} = |N_{i}^{r}|$, or can be automatically learned in a gradient based learning setup. 
$\sigma$ is an activation function such as ReLU, $W_{r}^{(1)}$ and $W_{0}^{(1)}$ are learnable parameters of the transformation.
\begin{table}[b]
\small
\begin{center}
 	\centering
	\begin{tabular}{cccC{2.5cm}}
		\toprule
		Relation & $p_s(u_i)$, $p_s(u_j)$ & $i<j$ & $(i, j)$ \\
		\midrule
		1 & $p_1$, $p_1$ & Yes & (1,3), (1,5), (3,5) \\
		2 & $p_1$, $p_1$ & No & (1,1), (3,1), (3,3) (5,1), (5,3), (5,5) \\
		3 & $p_2$, $p_2$ & Yes & (2,4)\\
		4 & $p_2$, $p_2$ & No & (2,2), (4,2), (4, 4)\\
		5 & $p_1$, $p_2$ & Yes & (1,2), (1,4), (3,4)\\
		6 & $p_1$, $p_2$ & No & (3,2), (5,2), (5,4)\\
		7 & $p_2$, $p_1$ & Yes & (2,3), (2,5), (4,5)\\
		8 & $p_2$, $p_1$ & No & (2,1), (4,1), (4,3)\\
		\bottomrule
	\end{tabular}
	\caption{$p_s(u_i)$ and $p_s(u_j)$ denotes the speaker of utterances $u_i$ and $u_j$. 2 distinct speakers in the conversation implies $2*M^2 = 2*2^2 = 8$ distinct relation types. 
	The rightmost column denotes the indices of the vertices of the constituting edge which has the relation type indicated by the leftmost column. \label{example}}
\end{center}
\end{table}
In the second step, another local neighbourhood based transformation is applied over the output of the first step,
\begin{equation}
\begin{split}
h_{i}^{(2)}  & = \sigma (\sum\limits_{j \in N_{i}^{r}}^{} W^{(2)} h_{j}^{(1)} + W_{0}^{(2)}h_{i}^{(1)}), \label{eq-ref2} \\
& \text{for }i = 1, 2,\dots, N,
\end{split}
\end{equation}
where,  $W^{(2)}$ and $W_{0}^{(2)}$ are parameters of these transformation and $\sigma$ is the activation function.

This stack of transformations, \cref{eq-ref1,eq-ref2}, effectively accumulates normalized sum of the local neighbourhood (features of the neighbours) i.e. the neighbourhood speaker information for each utterance in the graph. 
The self connection ensures self dependent feature transformation.

\paragraph{Emotion Classifier:} The contextually encoded feature vectors $g_i$ (from sequential encoder) and $h_{i}^{(2)}$ (from speaker-level encoder) are concatenated and a similarity-based attention mechanism is applied to obtain the final utterance representation:
\begin{flalign}
h_{i}  & = [g_{i}, h_{i}^{(2)}], \\
\beta_{i} & = \softmax(h_{i}^{T}W_{\beta}[h_{1},h_{2}\dots,h_{N}]), \label{eq:beta}\\
\tilde{h}_i&=\beta_i[h_1, h_2, \dots, h_N]^T. \label{eq:beta-2}
\end{flalign}
 Finally, the utterance is classified using a fully-connected network:
 \begin{flalign}
  l_i&=\ReLU(W_{l}\tilde{h}_i+b_{l}),\label{eq:5}\\
  \mathcal{P}_i&=\softmax(W_{smax}l_i+b_{smax}),\label{eq:c-6}\\
  \hat{y_i}&=\argmax_{k}(\mathcal{P}_i[k]).\label{eq:c-7}
\end{flalign}

\paragraph{Training Setup:}

We use categorical cross-entropy along with L2-regularization as the measure of loss ($L$) during training:
\begin{flalign}
  \label{eq:9}
  L=-\frac{1}{\sum_{s=1}^Nc(s)}\sum_{i=1}^N \sum_{j=1}^{c(i)}\log
  \mathcal{P}_{i,j}[y_{i,j}]+\lambda \norm{\theta}_2,
\end{flalign}
where $N$ is the number of samples/dialogues, $c(i)$ is the number of utterances in sample $i$, $\mathcal{P}_{i,j}$ is the probability distribution of emotion labels for utterance $j$ of dialogue $i$, $y_{i,j}$ is the expected class label of utterance $j$ of dialogue $i$, $\lambda$ is the L2-regularizer weight, and $\theta$ is the set of all trainable parameters. 

We used stochastic gradient descent based Adam~\cite{DBLP:journals/corr/KingmaB14} optimizer to
train our network. Hyperparameters were optimized using grid search. 

\section{Experimental Setting}
\label{sec:experiments}

\subsection{Datasets Used}
\label{sec:dataset-details}
We evaluate our DialogueGCN model on three benchmark datasets --- IEMOCAP \cite{iemocap}, AVEC \cite{Schuller:2012:ACA:2388676.2388776}, and MELD~\cite{poria2018meld}. All these three datasets are multimodal datasets containing textual, visual and acoustic information for every utterance of each conversation.
However, in this work we focus on conversational emotion recognition only from the textual information. Multimodal emotion recognition is outside the scope of this paper, and is left as future work.

\paragraph{IEMOCAP}~\cite{iemocap} dataset contains videos of two-way conversations of ten unique speakers, where only the first eight speakers from session one to four belong to the train-set. 
Each video contains a single dyadic dialogue, segmented into utterances. 
The utterances are annotated with one of six emotion labels, which are happy, sad, neutral, angry, excited, and frustrated.

\paragraph{AVEC}~\cite{Schuller:2012:ACA:2388676.2388776} dataset is a modification of
SEMAINE database~\cite{5959155} containing interactions between humans and
artificially intelligent agents. 
Each utterance of a dialogue is annotated with four real valued affective attributes: valence ($[-1,1]$), arousal ($[-1,1]$), expectancy ($[-1,1]$), and power ($[0,\infty)$). 
The annotations are available every 0.2 seconds in the original database. 
However, in order to adapt the annotations to our need of utterance-level annotation, we averaged the attributes over the span of an utterance.

\paragraph{MELD}~\cite{poria2018meld} is a multimodal emotion/sentiment classification dataset which has been created by the extending the EmotionLines dataset \cite{chen2018emotionlines}.
Contrary to IEMOCAP and AVEC, MELD is a multiparty dialog dataset.
MELD contains textual, acoustic and visual information for more than 1400 dialogues and 13000 utterances from the Friends TV series.
Each utterance in every dialog is annotated as one of the seven emotion classes: anger, disgust, sadness, joy, surprise, fear or neutral.

\begin{table}[ht!]
\centering
\resizebox{\linewidth}{!}{
        \begin{tabular}{l|c|c|c|c|c|c}
            \toprule
            \multirow{2}{*}{Dataset}&\multicolumn{3}{c|}{$\#$ dialogues}&\multicolumn{3}{c}{$\#$ utterances}\\
            &train&val&test&train&val&test\\
            \hline \hline
            IEMOCAP&\multicolumn{2}{c|}{120}&31&\multicolumn{2}{c|}{5810}&1623\\
            AVEC&\multicolumn{2}{c|}{63}&32&\multicolumn{2}{c|}{4368}&1430\\
            MELD&1039&114&280& 9989& 1109& 2610\\
            \bottomrule
        \end{tabular}
        }
    \caption{Training, validation and test data distribution in the datasets. No predefined train/val split is provided in IEMOCAP and AVEC, hence we use 10\% of the training dialogues as validation split.}
        \label{table:data1}
\end{table}



\subsection{Baselines and State of the Art}
\label{sec:baselines}

For a comprehensive evaluation of DialogueGCN, we compare our model with the following baseline methods:

\paragraph{CNN~\cite{kim2014convolutional}}
This is the baseline convolutional neural network based model which is identical to our utterance level feature extractor network (\cref{sec:text-feat-extr}). This model is context independent as it doesn't use information from contextual utterances.

\paragraph{Memnet~\cite{Sukhbaatar:2015:EMN:2969442.2969512}}
This is an end-to-end memory network baseline \cite{hazarika-EtAl:2018:N18-1}. Every utterance is fed to the network and the memories, which correspond to the previous utterances, is continuously updated in a multi-hop fashion. Finally the output from the memory network is used for emotion classification.

\paragraph{c-LSTM~\cite{poria-EtAl:2017:Long}}

Context-aware utterance representations are generated by capturing the contextual content from the surrounding utterances using a Bi-directional LSTM~\cite{hochreiter1997long} network. The context-aware utterance representations are then used for emotion classification. The contextual-LSTM model is speaker independent as it doesn't model any speaker level dependency.

\paragraph{c-LSTM+Att~\cite{poria-EtAl:2017:Long}}

In this variant of c-LSTM, an attention module is applied to the output of c-LSTM at each timestamp by following \cref{eq:beta,eq:beta-2}. Generally this provides better context to create a more informative final utterance representation.





\paragraph{CMN~\cite{hazarika-EtAl:2018:N18-1}}

CMN models utterance context from dialogue history
using two distinct GRUs for two speakers. 
Finally, utterance representation is obtained by feeding the current utterance as query to two distinct memory networks for both speakers. However, this model can only model conversations with two speakers. 

\paragraph{ICON~\cite{hazarika-EtAl:2018:N18-1}}
ICON which is an extension of CMN, connects outputs of individual speaker GRUs in CMN using another GRU for explicit inter-speaker modeling. This GRU is considered as a memory to track the overall conversational flow. Similar to CMN, ICON can not be extended to apply on multiparty datasets e.g., MELD.
\paragraph{DialogueRNN~\cite{dialoguernn}}

This is the state-of-the-art method for ERC. It is a recurrent network that uses two GRUs to track individual speaker states and global context during the conversation. Further, another GRU is employed to track emotional state through the conversation. DialogueRNN claims to model inter-speaker relation and it can be applied on multiparty datasets.

\section{Results and Discussions}
\label{sec:results}




\begin{table*}[t]
  \centering
  \resizebox{\linewidth}{!}{
   \begin{tabular}{l||c@{~~}c|c@{~~}c|c@{~~}c|c@{~~}c|c@{~~}c|c@{~~}c|c@{~~}c}
    \hline
    \multirow{3}{*}{Methods} & \multicolumn{14}{c}{IEMOCAP}  \\
    \cline{2-15} & \multicolumn{2}{c|}{Happy} & \multicolumn{2}{c|}{Sad} &
                                        \multicolumn{2}{c|}{Neutral} & \multicolumn{2}{c|}{Angry} & \multicolumn{2}{c|}{Excited} & \multicolumn{2}{c|}{Frustrated} & \multicolumn{2}{c}{\textbf{Average(w)}}\\
    \cline{2-15} & Acc. & F1 & Acc. & F1 & Acc. & F1 & Acc. & F1 & Acc. & F1 & Acc. & F1 & Acc. & F1 \\
    \hline
    \hline
  CNN &27.77&29.86&57.14&53.83&34.33&40.14&61.17&52.44&46.15&50.09&62.99&55.75&48.92&48.18 \\
    Memnet &25.72&33.53&55.53&61.77&58.12&52.84&59.32&55.39&51.50&58.30&67.20&59.00&55.72&55.10\\
    bc-LSTM &29.17&34.43&57.14&60.87&54.17&51.81&57.06&56.73&51.17&57.95&67.19&58.92&55.21&54.95\\
    bc-LSTM+Att &30.56&35.63&56.73&62.90&57.55&53.00&59.41&59.24&52.84&58.85&65.88&59.41&56.32&56.19\\
    CMN &25.00&30.38&55.92&62.41&52.86&52.39&61.76&59.83&55.52&60.25&71.13&60.69&56.56&56.13 \\
    ICON &22.22&29.91&58.78&64.57&62.76&57.38&64.71&63.04&58.86&63.42&67.19&60.81&59.09&58.54\\
    DialogueRNN &25.69&33.18&75.10&{78.80}&58.59&{59.21}&64.71&{\bf 65.28} &80.27&{\bf 71.86}&61.15&58.91&63.40&{62.75}\\
    \hline
    \textbf{DialogueGCN} &40.62 &{\bf 42.75} &89.14 &{\bf 84.54} &61.92 &{\bf 63.54} &67.53 &64.19 &65.46 &63.08 &64.18 &{\bf 66.99} &65.25 &{\bf 64.18} \\
    \hline
   \end{tabular}
  }
  \caption{Comparison with the baseline methods on IEMOCAP dataset; Acc. = Accuracy; bold font denotes the best performances. Average(w) = Weighted average.\label{results:IEMOCAP}}
\end{table*}


  
\subsection{Comparison with State of the Art and Baseline}
We compare the performance of our proposed DialogueGCN framework with the state-of-the-art DialogueRNN and baseline methods in \cref{results:IEMOCAP,results:AVEC}. We report all results with average of 5 runs. Our DialogueGCN model outperforms the SOTA and all the baseline models, on all the datasets, while also being statistically significant under the paired t-test (p \textless 0.05).

\paragraph{IEMOCAP and AVEC:} On the IEMOCAP dataset, DialogueGCN achieves new state-of-the-art average F1-score of 64.18\% and accuracy of 65.25\%, which is around 2\% better than DialogueRNN, and at least 5\% better than all the other baseline models. 
Similarly, on AVEC dataset, DialogueGCN outperforms the state-of-the-art on all the four emotion dimensions: valence, arousal, expectancy, and power.

To explain this gap in performance, it is important to understand the nature of these models. DialogueGCN and DialogueRNN both try to model speaker-level context (albeit differently), whereas, none of the other models encode speaker-level context (they only encode sequential context). This is a key limitation in the baseline models, as speaker-level context is indeed very important in conversational emotion recognition. 

As for the difference of performance between DialogueRNN and DialogueGCN, we believe that this is due to the different nature of speaker-level context encoding. DialogueRNN employs a gated recurrent unit (GRU) network to model individual speaker states. Both IEMOCAP and AVEC dataset has many conversations with over 70 utterances (the average conversation length is 50 utterances in IEMOCAP and 72 in AVEC). As recurrent encoders have long-term information propagation issues, speaker-level encoding can be problematic for long sequences like those found in these two datasets. In contrast, DialogueGCN tries to overcome this issue by using neighbourhood based convolution to model speaker-level context. 
\begin{figure*}[t]
    \centering
    \begin{subfigure}[t]{0.47\textwidth}
        \centering
        \includegraphics[width=\linewidth]{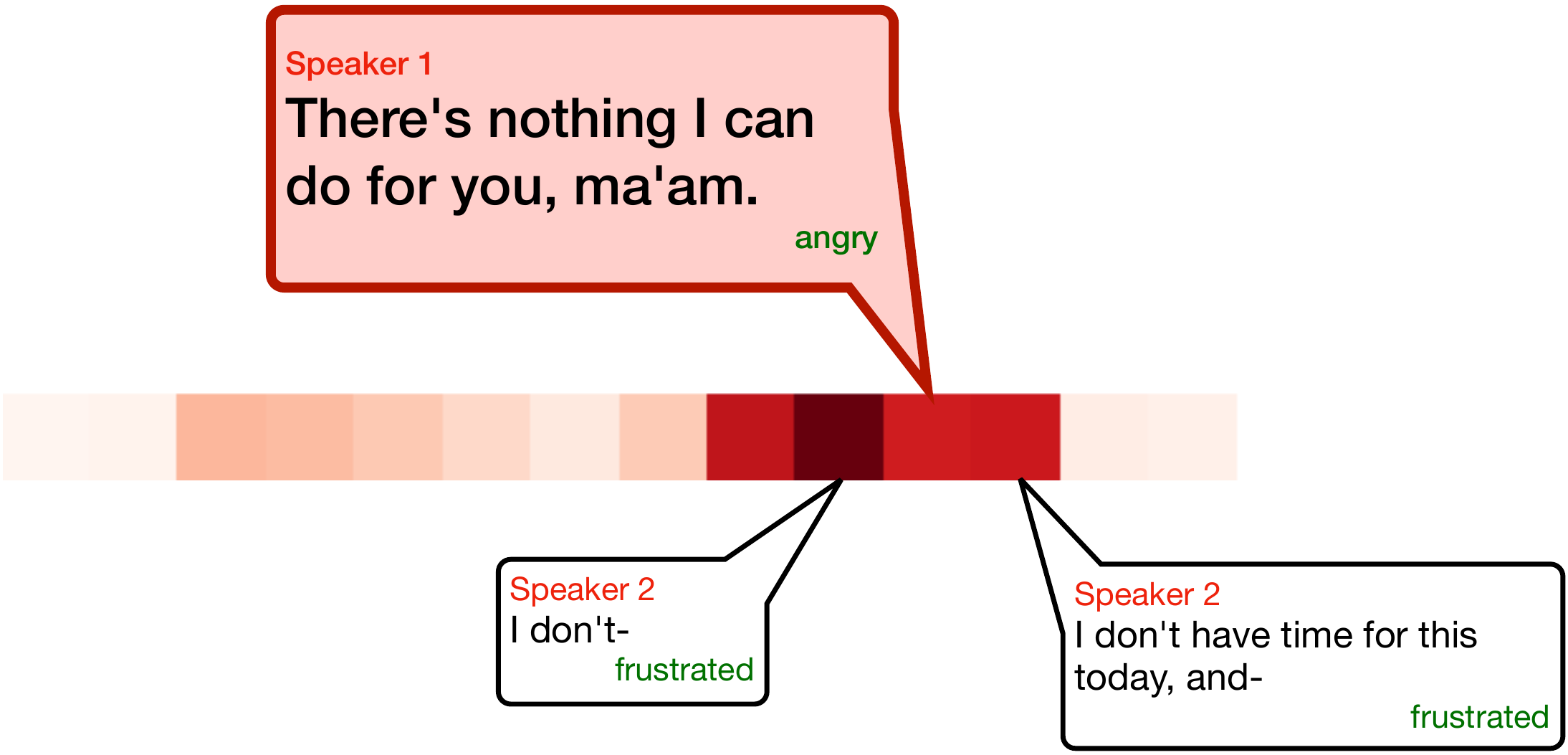}
        \caption{}
        \label{fig:dgcn_inters}
    \end{subfigure}
    ~
    \begin{subfigure}[t]{0.47\textwidth}
        \centering
        \includegraphics[width=\linewidth]{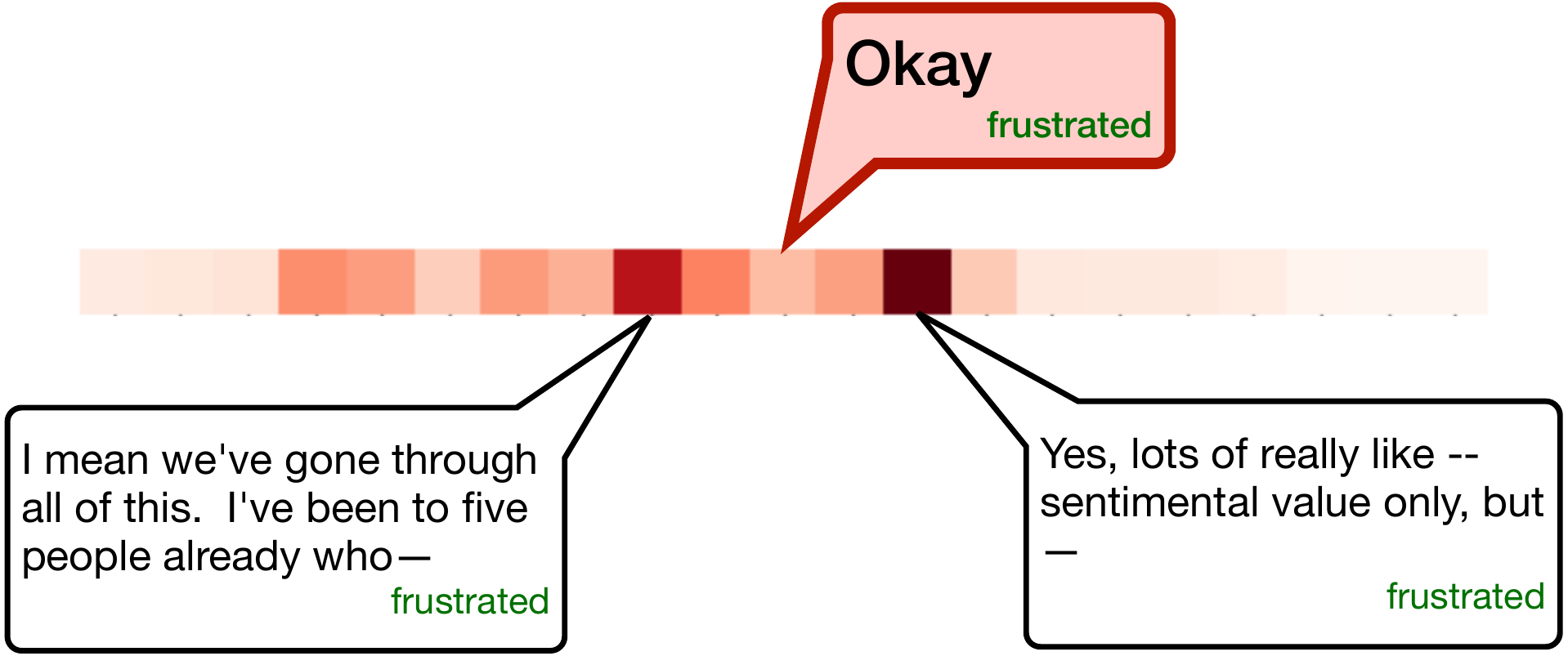}
        \caption{}
        \label{fig:dgcn_short}
    \end{subfigure}
        \caption{Visualization of edge-weights in \cref{eq:edge_weights} --- (a) Target utterance attends to other speaker's utterance for correct context; (b) Short utterance attends to appropriate contextual utterances to be classified correctly. }
\end{figure*}
\paragraph{MELD:} The MELD dataset consists of multiparty conversations and we found that emotion recognition in MELD is considerably harder to model than IEMOCAP and AVEC - which only consists of dyadic conversations. Utterances in MELD are much shorter and rarely contain emotion specific expressions, which means emotion modelling is highly context dependent. Moreover, the average conversation length is 10 utterances, with many conversations having more than 5 participants, which means majority of the participants only utter a small number of utterances per conversation. This makes inter-dependency and self-dependency modeling difficult.
Because of these reasons, we found that the difference in results between the baseline models and DialogueGCN is not as contrasting as it is in the case of IEMOCAP and AVEC. Memnet, CMN, and ICON are not suitable for this dataset as they exclusively work in dyadic conversations. Our DialogueGCN model achieves new state-of-the-art F1 score of 58.10\% outperforming DialogueRNN by more than 1\%. We surmise that this improvement is a result of the speaker dependent relation modelling of the edges in our graph network which inherently improves the context understanding over DialogueRNN.
\begin{table*}[t]
\centering
   \begin{tabular}{l||c@{~}c@{~}c@{~}c@{~}|c}
    \hline
    \multirow{3}{*}{Methods} & \multicolumn{4}{c|}{AVEC} & \multirow{2}{*}{MELD}\\
    \cline{2-5} &\multicolumn{1}{c}{Valence}& \multicolumn{1}{c}{Arousal}& \multicolumn{1}{c}{Expectancy} & \multicolumn{1}{c|}{Power}\\
    \hline
    \hline
  CNN &0.545&0.542&0.605&8.71 & 55.02 \\
    Memnet &0.202 &0.211&0.216&8.97 & -\\
    bc-LSTM &0.194&0.212&0.201&8.90 & 56.44\\
    bc-LSTM+Att &0.189&0.213&0.190&8.67 & 56.70\\
    CMN &0.192&0.213&0.195&8.74 & -\\
    ICON & 0.180&0.190&0.180&8.45 & -\\
    DialogueRNN &0.168&0.165& 0.175&7.90 & 57.03\\
    \hline
    \textbf{DialogueGCN} &{\bf 0.157}&{\bf 0.161}&{\bf 0.168}&{\bf 7.68} & {\bf 58.10}\\
    \hline
   \end{tabular}
  \caption{Comparison with the baseline methods on AVEC and MELD dataset; MAE and F1 metrics are user for AVEC and MELD, respectively.\label{results:AVEC}}
  \end{table*}
\subsection{Effect of Context Window}
We report results for DialogueGCN model in \cref{results:IEMOCAP,results:AVEC} with a past and future context window size of (10, 10) to construct the edges. We also carried out experiments with decreasing context window sizes of (8, 8), (4, 4), (0, 0) and found that performance steadily decreased with F1 scores of 62.48\%, 59.41\% and 55.80\% on IEMOCAP. DialogueGCN with context window size of (0, 0) is equivalent to a model with only sequential encoder (as it only has self edges), and performance is expectedly much worse. 
We couldn't perform extensive experiments with larger windows because of computational constraints, but we expect performance to improve with larger context sizes.
\subsection{Ablation Study}
We perform ablation study for different level of context encoders, namely sequential encoder and speaker-level encoder, in \cref{table:abl1}. We remove them one at a time and found that the speaker-level encoder is slightly more important in overall performance. This is due to speaker-level encoder mitigating long distance dependency issue of sequential encoder and DialogueRNN. Removing both of them results in a very poor F1 score of 36.7 \%, which demonstrates the importance of contextual modelling in conversational emotion recognition.
\begin{table}[ht!]
 	\centering
	\begin{tabular}{C{2cm}C{2.5cm}c}
		\toprule
		Sequential Encoder & Speaker-Level Encoder & F1  \\
		\midrule
		\cmark & \cmark &  64.18 \\
		\cmark & \xmark &  55.30 \\
		\xmark & \cmark &  56.71 \\
		\xmark & \xmark &  36.75 \\
		\bottomrule
	\end{tabular}
	\caption{Ablation results w.r.t the contextual encoder modules on IEMOCAP dataset.\label{table:abl1}}
\end{table}

\begin{table}[ht!]
 	\centering
	\begin{tabular}{C{2.5cm}C{2.5cm}c}
		\toprule
		Speaker Dependency Edges & Temporal Dependency Edges & F1  \\
		\midrule
		\cmark & \cmark &  64.18 \\
		\cmark & \xmark &  62.52 \\
		\xmark & \cmark &  61.03 \\
		\xmark & \xmark &  60.11 \\
		\bottomrule
	\end{tabular}
	\caption{Ablation results w.r.t the edge relations in speaker-level encoder module on IEMOCAP dataset. \label{table:abl2}}
\end{table}

Further, we study the effect of edge relation modelling. As mentioned in \cref{sec:spealer-level}, there are total $2M^2$ distinct edge relations 
for a conversation with $M$ distinct speakers. First we removed only the temporal dependency (resulting in $M^2$ distinct edge relations), and then only the speaker dependency (resulting in $2$ distinct edge relations) and then both (resulting in a single edge relation all throughout the graph). The results of these tests in \cref{table:abl2} show that having these different relational edges is indeed very important for modelling emotional dynamics. These results support our hypothesis that each speaker in a conversation is uniquely affected by the others, and hence, modelling interlocutors-dependency is rudimentary. \cref{fig:dgcn_inters} illustrates one such instance where target utterance attends to other speaker's utterance for context. This phenomenon is commonly observable for DialogueGCN, as compared to DialogueRNN.

\subsection{Performance on Short Utterances}
Emotion of short utterances, like ``okay'', ``yeah'', depends on the context it appears in. For example, without context ``okay'' is assumed `neutral'. However, in \cref{fig:dgcn_short}, DialogueGCN correctly classifies ``okay'' as `frustration', which is apparent from the context. We observed that, overall, DialogueGCN correctly classifies short utterances, where DialogueRNN fails.

\subsection{Error Analysis}
We analyzed our predicted emotion labels and found that misclassifications are often among similar emotion classes. In the confusion matrix, we observed that our model misclassifies several samples of `frustrated' as `angry' and `neutral'. This is due to subtle difference between frustration and anger. Further, we also observed similar misclassification of `excited' samples as `happy' and `neutral'. All the datasets that we use in our experiment are multimodal. A few utterances e.g., `ok. yes' carrying \emph{non-neutral} emotions were misclassified as we do not utilize audio and visual modality in our experiments. In such utterances, we found audio and visual (in this particular example, high pitched audio and frowning expression) modality providing key information to detect underlying emotions (\emph{frustrated} in the above utterance) which DialogueGCN failed to understand by just looking at the textual context.

\section{Conclusion}
\label{sec:conclusion}
In this work, we present Dialogue Graph Convolutional Network (DialogueGCN), that models inter and self-party dependency to improve context understanding for utterance-level emotion detection in conversations. On three benchmark ERC datasets, DialogueGCN outperforms the strong baselines and existing state of the art, by a significant margin. Future works will focus on incorporating multimodal information into DialogueGCN, speaker-level emotion shift detection, and conceptual grounding of conversational emotion reasoning. We also plan to use DialogueGCN in dialogue systems to generate affective responses.
\bibliography{emnlp-ijcnlp-2019}

\begin{thebibliography}{34}
\expandafter\ifx\csname natexlab\endcsname\relax\def\natexlab#1{#1}\fi

\bibitem[{Bradbury et~al.(2017)Bradbury, Merity, Xiong, and
  Socher}]{bradbury2016quasi}
James Bradbury, Stephen Merity, Caiming Xiong, and Richard Socher. 2017.
\newblock {Quasi-Recurrent Neural Networks}.
\newblock In \emph{International Conference on Learning Representations (ICLR
  2017)}.

\bibitem[{Bruna et~al.(2013)Bruna, Zaremba, Szlam, and
  LeCun}]{bruna2013spectral}
Joan Bruna, Wojciech Zaremba, Arthur Szlam, and Yann LeCun. 2013.
\newblock Spectral networks and locally connected networks on graphs.
\newblock \emph{arXiv preprint arXiv:1312.6203}.

\bibitem[{Busso et~al.(2008)Busso, Bulut, Lee, Kazemzadeh, Mower, Kim, Chang,
  Lee, and Narayanan}]{iemocap}
Carlos Busso, Murtaza Bulut, Chi-Chun Lee, Abe Kazemzadeh, Emily Mower, Samuel
  Kim, Jeannette~N Chang, Sungbok Lee, and Shrikanth~S Narayanan. 2008.
\newblock {IEMOCAP: Interactive emotional dyadic motion capture database}.
\newblock \emph{Language resources and evaluation}, 42(4):335--359.

\bibitem[{Chen et~al.(2017)Chen, Wang, Liang, Baltru{\v{s}}aitis, Zadeh, and
  Morency}]{chen2017multimodal}
Minghai Chen, Sen Wang, Paul~Pu Liang, Tadas Baltru{\v{s}}aitis, Amir Zadeh,
  and Louis-Philippe Morency. 2017.
\newblock Multimodal sentiment analysis with word-level fusion and
  reinforcement learning.
\newblock In \emph{Proceedings of the 19th ACM International Conference on
  Multimodal Interaction}, pages 163--171. ACM.

\bibitem[{Chen et~al.(2018)Chen, Hsu, Kuo, Ku et~al.}]{chen2018emotionlines}
Sheng-Yeh Chen, Chao-Chun Hsu, Chuan-Chun Kuo, Lun-Wei Ku, et~al. 2018.
\newblock Emotionlines: An emotion corpus of multi-party conversations.
\newblock \emph{arXiv preprint arXiv:1802.08379}.

\bibitem[{Chung et~al.(2014)Chung, G{\"{u}}l{\c{c}}ehre, Cho, and Bengio}]{gru}
Junyoung Chung, {\c{C}}aglar G{\"{u}}l{\c{c}}ehre, KyungHyun Cho, and Yoshua
  Bengio. 2014.
\newblock \href {http://arxiv.org/abs/1412.3555} {{Empirical Evaluation of
  Gated Recurrent Neural Networks on Sequence Modeling}}.
\newblock \emph{CoRR}, abs/1412.3555.

\bibitem[{Colneri{\^c} and Demsar(2018)}]{colneric2018emotion}
Niko Colneri{\^c} and Janez Demsar. 2018.
\newblock Emotion recognition on twitter: comparative study and training a
  unison model.
\newblock \emph{IEEE Transactions on Affective Computing}.

\bibitem[{Defferrard et~al.(2016)Defferrard, Bresson, and
  Vandergheynst}]{NIPS2016_6081}
Micha\"{e}l Defferrard, Xavier Bresson, and Pierre Vandergheynst. 2016.
\newblock \href
  {http://papers.nips.cc/paper/6081-convolutional-neural-networks-on-graphs-with-fast-localized-spectral-filtering.pdf}
  {{Convolutional Neural Networks on Graphs with Fast Localized Spectral
  Filtering}}.
\newblock In D.~D. Lee, M.~Sugiyama, U.~V. Luxburg, I.~Guyon, and R.~Garnett,
  editors, \emph{Advances in Neural Information Processing Systems 29}, pages
  3844--3852. Curran Associates, Inc.

\bibitem[{Gilmer et~al.(2017)Gilmer, Schoenholz, Riley, Vinyals, and
  Dahl}]{gilmer2017neural}
Justin Gilmer, Samuel~S Schoenholz, Patrick~F Riley, Oriol Vinyals, and
  George~E Dahl. 2017.
\newblock Neural message passing for quantum chemistry.
\newblock In \emph{Proceedings of the 34th International Conference on Machine
  Learning-Volume 70}, pages 1263--1272. JMLR. org.

\bibitem[{Hazarika et~al.(2018{\natexlab{a}})Hazarika, Poria, Mihalcea,
  Cambria, and Zimmermann}]{hazarika2018icon}
Devamanyu Hazarika, Soujanya Poria, Rada Mihalcea, Erik Cambria, and Roger
  Zimmermann. 2018{\natexlab{a}}.
\newblock Icon: Interactive conversational memory network for multimodal
  emotion detection.
\newblock In \emph{Proceedings of the 2018 Conference on Empirical Methods in
  Natural Language Processing}, pages 2594--2604.

\bibitem[{Hazarika et~al.(2018{\natexlab{b}})Hazarika, Poria, Zadeh, Cambria,
  Morency, and Zimmermann}]{hazarika-EtAl:2018:N18-1}
Devamanyu Hazarika, Soujanya Poria, Amir Zadeh, Erik Cambria, Louis-Philippe
  Morency, and Roger Zimmermann. 2018{\natexlab{b}}.
\newblock \href {http://www.aclweb.org/anthology/N18-1193} {{Conversational
  Memory Network for Emotion Recognition in Dyadic Dialogue Videos}}.
\newblock In \emph{Proceedings of the 2018 Conference of the North American
  Chapter of the Association for Computational Linguistics: Human Language
  Technologies, Volume 1 (Long Papers)}, pages 2122--2132, New Orleans,
  Louisiana. Association for Computational Linguistics.

\bibitem[{Hochreiter and Schmidhuber(1997)}]{hochreiter1997long}
Sepp Hochreiter and J{\"u}rgen Schmidhuber. 1997.
\newblock Long short-term memory.
\newblock \emph{Neural computation}, 9(8):1735--1780.

\bibitem[{Huang et~al.(2019)Huang, Trabelsi, and Za{\"\i}ane}]{huang2019ana}
Chenyang Huang, Amine Trabelsi, and Osmar~R Za{\"\i}ane. 2019.
\newblock Ana at semeval-2019 task 3: Contextual emotion detection in
  conversations through hierarchical lstms and bert.
\newblock \emph{arXiv preprint arXiv:1904.00132}.

\bibitem[{K.~D'Mello et~al.(2006)K.~D'Mello, Craig, Sullins, and
  Graesser}]{dmello}
Sidney K.~D'Mello, Scotty Craig, Jeremiah Sullins, and Arthur Graesser. 2006.
\newblock Predicting affective states expressed through an emote-aloud
  procedure from autotutor's mixed-initiative dialogue.
\newblock \emph{I. J. Artificial Intelligence in Education}, 16:3--28.

\bibitem[{Kim(2014)}]{kim2014convolutional}
Yoon Kim. 2014.
\newblock Convolutional neural networks for sentence classification.
\newblock In \emph{EMNLP 2014}, pages 1746--1751.

\bibitem[{Kingma and Ba(2014)}]{DBLP:journals/corr/KingmaB14}
Diederik~P. Kingma and Jimmy Ba. 2014.
\newblock \href {http://arxiv.org/abs/1412.6980} {{Adam: A Method for
  Stochastic Optimization}}.
\newblock \emph{CoRR}, abs/1412.6980.

\bibitem[{Kipf and Welling(2016)}]{kipf2016semi}
Thomas~N Kipf and Max Welling. 2016.
\newblock Semi-supervised classification with graph convolutional networks.
\newblock \emph{arXiv preprint arXiv:1609.02907}.

\bibitem[{Kratzwald et~al.(2018)Kratzwald, Ilic, Kraus, Feuerriegel, and
  Prendinger}]{kratzwald2018decision}
Bernhard Kratzwald, Suzana Ilic, Mathias Kraus, Stefan Feuerriegel, and Helmut
  Prendinger. 2018.
\newblock Decision support with text-based emotion recognition: Deep learning
  for affective computing.
\newblock \emph{arXiv preprint arXiv:1803.06397}.

\bibitem[{Majumder et~al.(2019)Majumder, Poria, Hazarika, Mihalcea, Gelbukh,
  and Cambria}]{dialoguernn}
Navonil Majumder, Soujanya Poria, Devamanyu Hazarika, Rada Mihalcea, Alexander
  Gelbukh, and Erik Cambria. 2019.
\newblock \href {https://doi.org/10.1609/aaai.v33i01.33016818} {{DialogueRNN:
  An Attentive RNN for Emotion Detection in Conversations}}.
\newblock In \emph{Proceedings of the AAAI Conference on Artificial
  Intelligence}, volume~33, pages 6818--6825.

\bibitem[{McKeown et~al.(2012)McKeown, Valstar, Cowie, Pantic, and
  Schroder}]{5959155}
G.~McKeown, M.~Valstar, R.~Cowie, M.~Pantic, and M.~Schroder. 2012.
\newblock \href {https://doi.org/10.1109/T-AFFC.2011.20} {{The SEMAINE
  Database: Annotated Multimodal Records of Emotionally Colored Conversations
  between a Person and a Limited Agent}}.
\newblock \emph{IEEE Transactions on Affective Computing}, 3(1):5--17.

\bibitem[{Nair and Hinton(2010)}]{nair2010rectified}
Vinod Nair and Geoffrey~E Hinton. 2010.
\newblock Rectified linear units improve restricted boltzmann machines.
\newblock In \emph{Proceedings of the 27th international conference on machine
  learning (ICML-10)}, pages 807--814.

\bibitem[{Navarretta et~al.(2016)Navarretta, Choukri, Declerck, Goggi,
  Grobelnik, and Maegaard}]{navarretta2016mirroring}
Costanza Navarretta, K~Choukri, T~Declerck, S~Goggi, M~Grobelnik, and
  B~Maegaard. 2016.
\newblock Mirroring facial expressions and emotions in dyadic conversations.
\newblock In \emph{LREC}.

\bibitem[{Pennington et~al.(2014)Pennington, Socher, and
  Manning}]{pennington2014glove}
Jeffrey Pennington, Richard Socher, and Christopher~D. Manning. 2014.
\newblock \href {http://www.aclweb.org/anthology/D14-1162} {Glove: Global
  vectors for word representation}.
\newblock In \emph{Empirical Methods in Natural Language Processing (EMNLP)},
  pages 1532--1543.

\bibitem[{Poria et~al.(2017)Poria, Cambria, Hazarika, Majumder, Zadeh, and
  Morency}]{poria-EtAl:2017:Long}
Soujanya Poria, Erik Cambria, Devamanyu Hazarika, Navonil Majumder, Amir Zadeh,
  and Louis-Philippe Morency. 2017.
\newblock \href {http://aclweb.org/anthology/P17-1081} {{Context-Dependent
  Sentiment Analysis in User-Generated Videos}}.
\newblock In \emph{Proceedings of the 55th Annual Meeting of the Association
  for Computational Linguistics (Volume 1: Long Papers)}, pages 873--883,
  Vancouver, Canada. Association for Computational Linguistics.

\bibitem[{Poria et~al.(2019{\natexlab{a}})Poria, Hazarika, Majumder, Naik,
  Cambria, and Mihalcea}]{poria2018meld}
Soujanya Poria, Devamanyu Hazarika, Navonil Majumder, Gautam Naik, Erik
  Cambria, and Rada Mihalcea. 2019{\natexlab{a}}.
\newblock \href {https://www.aclweb.org/anthology/P19-1050} {{MELD}: A
  multimodal multi-party dataset for emotion recognition in conversations}.
\newblock In \emph{Proceedings of the 57th Annual Meeting of the Association
  for Computational Linguistics}, pages 527--536, Florence, Italy. Association
  for Computational Linguistics.

\bibitem[{Poria et~al.(2019{\natexlab{b}})Poria, Majumder, Mihalcea, and
  Hovy}]{poria2019emotion}
Soujanya Poria, Navonil Majumder, Rada Mihalcea, and Eduard Hovy.
  2019{\natexlab{b}}.
\newblock \href {https://doi.org/10.1109/ACCESS.2019.2929050} {Emotion
  recognition in conversation: Research challenges, datasets, and recent
  advances}.
\newblock \emph{IEEE Access}, 7:100943--100953.

\bibitem[{Scarselli et~al.(2008)Scarselli, Gori, Tsoi, Hagenbuchner, and
  Monfardini}]{scarselli2008graph}
Franco Scarselli, Marco Gori, Ah~Chung Tsoi, Markus Hagenbuchner, and Gabriele
  Monfardini. 2008.
\newblock The graph neural network model.
\newblock \emph{IEEE Transactions on Neural Networks}, 20(1):61--80.

\bibitem[{Schlichtkrull et~al.(2018)Schlichtkrull, Kipf, Bloem, Van Den~Berg,
  Titov, and Welling}]{schlichtkrull2018modeling}
Michael Schlichtkrull, Thomas~N Kipf, Peter Bloem, Rianne Van Den~Berg, Ivan
  Titov, and Max Welling. 2018.
\newblock Modeling relational data with graph convolutional networks.
\newblock In \emph{European Semantic Web Conference}, pages 593--607. Springer.

\bibitem[{Schuller et~al.(2012)Schuller, Valster, Eyben, Cowie, and
  Pantic}]{Schuller:2012:ACA:2388676.2388776}
Bj\"{o}rn Schuller, Michel Valster, Florian Eyben, Roddy Cowie, and Maja
  Pantic. 2012.
\newblock \href {https://doi.org/10.1145/2388676.2388776} {{AVEC 2012: The
  Continuous Audio/Visual Emotion Challenge}}.
\newblock In \emph{Proceedings of the 14th ACM International Conference on
  Multimodal Interaction}, ICMI '12, pages 449--456, New York, NY, USA. ACM.

\bibitem[{Strapparava and Mihalcea(2010)}]{strapparava2010annotating}
Carlo Strapparava and Rada Mihalcea. 2010.
\newblock Annotating and identifying emotions in text.
\newblock In \emph{Intelligent Information Access}, pages 21--38. Springer.

\bibitem[{Sukhbaatar et~al.(2015)Sukhbaatar, Szlam, Weston, and
  Fergus}]{Sukhbaatar:2015:EMN:2969442.2969512}
Sainbayar Sukhbaatar, Arthur Szlam, Jason Weston, and Rob Fergus. 2015.
\newblock \href {http://dl.acm.org/citation.cfm?id=2969442.2969512}
  {{End-to-end Memory Networks}}.
\newblock In \emph{Proceedings of the 28th International Conference on Neural
  Information Processing Systems - Volume 2}, NIPS'15, pages 2440--2448,
  Cambridge, MA, USA. MIT Press.

\bibitem[{Zadeh et~al.(2018{\natexlab{a}})Zadeh, Liang, Mazumder, Poria,
  Cambria, and Morency}]{AAAI1817341}
Amir Zadeh, Paul~Pu Liang, Navonil Mazumder, Soujanya Poria, Erik Cambria, and
  Louis-Philippe Morency. 2018{\natexlab{a}}.
\newblock \href
  {https://aaai.org/ocs/index.php/AAAI/AAAI18/paper/view/17341/16122} {{Memory
  Fusion Network for Multi-view Sequential Learning}}.
\newblock In \emph{AAAI Conference on Artificial Intelligence}, pages
  5634--5641.

\bibitem[{Zadeh et~al.(2018{\natexlab{b}})Zadeh, Liang, Poria, Vij, Cambria,
  and Morency}]{zadatt}
Amir Zadeh, Paul~Pu Liang, Soujanya Poria, Prateek Vij, Erik Cambria, and
  Louis-Philippe Morency. 2018{\natexlab{b}}.
\newblock Multi-attention recurrent network for human communication
  comprehension.
\newblock In \emph{{Proceedings of the AAAI Conference on Artificial
  Intelligence}}, pages 5642--5649.

\bibitem[{Zhou et~al.(2018)Zhou, Huang, Zhang, Zhu, and
  Liu}]{zhou2018emotional}
Hao Zhou, Minlie Huang, Tianyang Zhang, Xiaoyan Zhu, and Bing Liu. 2018.
\newblock Emotional chatting machine: Emotional conversation generation with
  internal and external memory.
\newblock In \emph{Thirty-Second AAAI Conference on Artificial Intelligence}.

\end{thebibliography}
\bibliographystyle{acl_natbib}
\appendix
\end{document}